\documentclass{article}
\usepackage{spconf,amsmath,epsfig}
\usepackage[pagebackref,breaklinks,colorlinks]{hyperref}
\usepackage{amssymb}
\usepackage{threeparttable}
\usepackage{multirow} % multirow for format of table
\usepackage{subcaption} % subfigure

\newcommand{\etal}{\emph{et~al.~}}
\newcommand{\ie}{\emph{i.e.}}

\newcommand{\etc}{\emph{etc}}

\title{CONTEXTUAL ADVERSARIAL ATTACK AGAINST AERIAL DETECTION\\ IN THE PHYSICAL WORLD}
\name{Jiawei Lian\textsuperscript{1}, Xiaofei Wang\textsuperscript{1}, Yuru Su\textsuperscript{1}, Mingyang Ma\textsuperscript{2}, Shaohui Mei\textsuperscript{1*}\thanks{This work was supported by the National Natural Science Foundation of China (62171381 and 62271409). Corresponding author: Shaohui Mei (meish@nwpu.edu.cn).}}
\address{\textsuperscript{1}School of Electronics and Information, Northwestern Polytechnical University, Xi’an 710129, China\\
\textsuperscript{2}School of Information and Communications Engineering, Xi'an Jiaotong University, Xi’an 710049, China
}
\begin{document}
%\ninept
%
\maketitle
\begin{abstract}
Deep Neural Networks (DNNs) have been extensively utilized in aerial detection. However, DNNs' sensitivity and vulnerability to maliciously elaborated adversarial examples have progressively garnered attention. 
Recently, physical attacks have gradually become a hot issue due to they are more practical in the real world, which poses great threats to some security-critical applications.
In this paper, we take the first attempt to perform physical attacks in contextual form against aerial detection in the physical world.
We propose an innovative contextual attack method against aerial detection in real scenarios, which achieves powerful attack performance and transfers well between various aerial object detectors without smearing or blocking the interested objects to hide.
Based on the findings that the targets' contextual information plays an important role in aerial detection by observing the detectors' attention maps, we propose to make full use of the contextual area of the interested targets to elaborate contextual perturbations for the uncovered attacks in real scenarios.
Extensive proportionally scaled experiments are conducted to evaluate the effectiveness of the proposed contextual attack method, which demonstrates the proposed method's superiority in both attack efficacy and physical practicality.
\end{abstract}
\begin{keywords}
Adversarial examples, contextual perturbations, physical attacks, aerial detection
\end{keywords}
\section{INTRODUCTION}
\label{sec:introduction}

In recent years, deep learning technology based on Deep Neural Networks (DNNs) has made great breakthroughs in computer vision, and natural language processing.
Thus, DNNs have been widely applied in business and industry, such as mobile payment, autonomous driving, medical diagnosis, intelligent security, robotics, and other fields. 

However, the widespread application of DNNs also buries potential safety hazards. 
Szegedy \etal \cite{szegedy2014intriguing} first designed an adversarial perturbation imperceptible to humans and added it to clean images to generate adversarial examples, which can misguide DNNs make completely different wrong predictions. 
Such malicious behavior and maliciously designed examples are named adversarial attacks and adversarial examples, respectively, and the attacked model is also called the victim model. 
Since then, various deep learning tasks have fallen under adversarial attacks, such as image classification, object detection, spam detection, malware identification, natural language processing, deep reinforcement learning, \etc. 
All DNNs-based models show great sensitivity and vulnerability in the face of adversarial examples. 

Computer vision tasks, according to different attack domains, can be divided into digital attacks and physical attacks. 
Digital attacks refer to attack by tampering with the image pixels in the digital domain after imaging, while physical attacks refer to attack by tampering with the interested targets before imaging. 
Digital attack methods can easily fool various deep learning models in the digital domain.
Since the generated digital perturbations typically cover the entire image and are invisible to humans, making them uncapturable by imaging devices. 
This problem drives more scholars to delve into the adversarial attacks applicable to real scenarios. 
Consequently, many physical attacks in patch form have been proposed to deceive intelligent systems such as autonomous driving \cite{wang2021dual}, face recognition \cite{wei2022adversarial}, and aerial detection \cite{lian2022benchmarking} in real-world scenarios.

In this work, we devote ourselves to conducting contextual attacks (CA) against aerial detection in physical world scenarios. The main contributions are summarized as follows:
\begin{itemize}
  \item We propose a novel contextual attack against aerial detection in physical scenarios, which achieves strong attack efficacy in both white-box and black-box conditions without smudging or blocking the targets to hide.
  \item We find that the targets' context information plays a key role in detection by observing their attention maps. 
  Thus we make full use of the contextual feature of the interested targets to elaborate contextual perturbations.
  \item We evaluate the proposed contextual attack method with two SOTA methods by performing proportionally scaled experiments, demonstrating our method's superiority in both attack efficacy and physical robustness.
\end{itemize}

\begin{figure*}[!t]
\centering
\includegraphics*[width=0.98\linewidth]{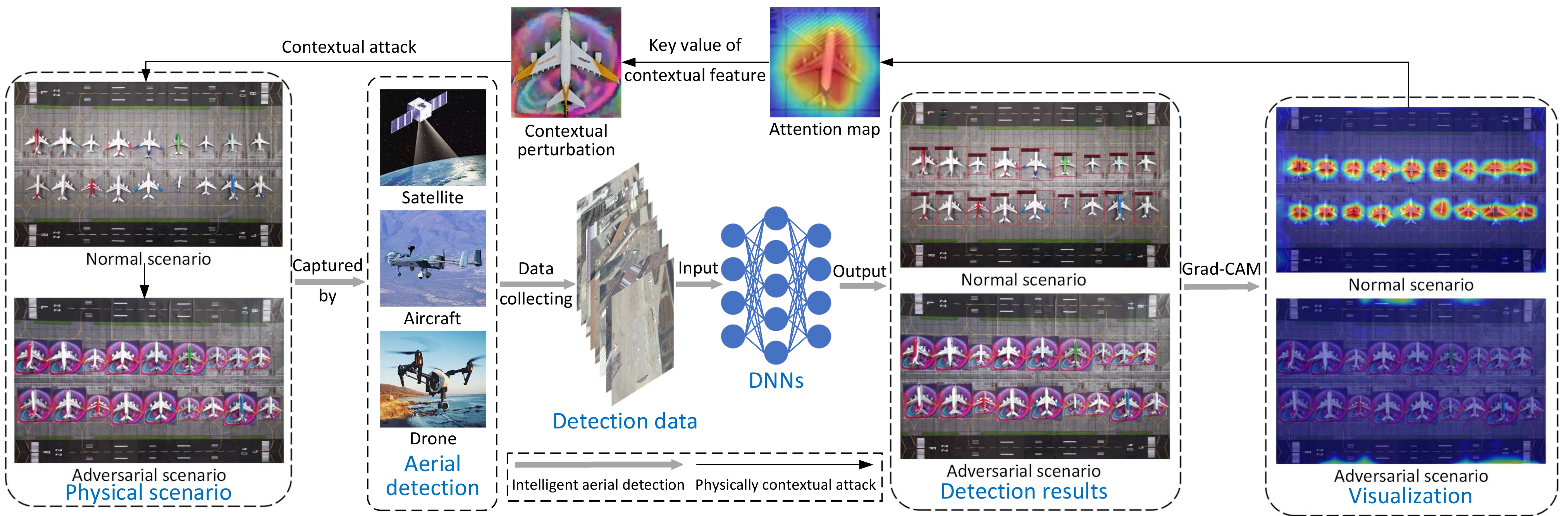}
\caption{
The illustration of the proposed contextual attack against aerial detection in physical scenarios.
%Firstly, we extract the aircraft mask to separate the aircraft area from the adversarial patch. 
%Secondly, augmentations are adopted to adapt the physical dynamics. 
%Thirdly, we place the adversarial patches on the clean image in the proper size and location to generate an adversarial example.
%Then, the adversarial example is fed into an aerial detector, and the recognized targets are used to optimize the adversarial patch.
%Finally, we use the background pixels of the adversarial patch plus the extracted aircraft to update the adversarial aircraft, \ie, we only optimize the background pixels of the adversarial aircraft during the training process.
}
\label{fig:Overview}
\end{figure*}

\section{METHODOLOGY}
\label{sec:methodology}

\subsection{Problem Formulation}

In this work, we aim to design contextual perturbations in patch form for unblocked attacks in the physical world.
Given a clean example $\boldsymbol{x}$, by attaching the contextual perturbations on clean image $\boldsymbol{x}$ to generate adversarial example $\boldsymbol{x}^*$.
Technically, the adversarial example is formulated as follows:
\begin{equation}
    \label{eq_problem-formulation}
    \boldsymbol{x}^* = (1-\boldsymbol{M}_{\boldsymbol{P}^*}) \odot \boldsymbol{x} + \boldsymbol{M}_{\boldsymbol{P}^*} \odot \boldsymbol{P}^*,
\end{equation}
where $\odot$ and $\boldsymbol{P}^*$ represent Hadamard product and contextual perturbations, respectively. 
Perturbations' mask $\boldsymbol{M}_{\boldsymbol{P}^*}$ is applied to properly attach contextual perturbations on the interested targets of a benign example.

\subsection{Overall Framework}

The overall framework of the proposed contextual attack is shown in Fig. \ref{fig:Overview}.
In the physical world, normal scenarios are captured by various remote sensing devices, such as satellites, aircraft, and drones.
Then, various DNNs-based aerial detectors are adopted to process massive aerial imagery.
To better understand the DNNs' predictions, we use Grad-CAM \cite{selvaraju2017grad} to visualize the aerial detectors' attention maps.
It is observed that the detectors also focus on the targets' contextual area beside the targets themselves.
Based on the findings that the targets' contextual information plays an important role in aerial detection by observing the aerial detectors' attention maps.
Thus we propose to fully use the contextual area of the interested targets to elaborate contextual perturbations for the uncovered attacks in real scenarios.
Finally, the elaborately designed contextual perturbations are applied in the physical world to hide targeted objects from being recognized.

\begin{table*}[t!]
\footnotesize
\caption{Quantitative experimental results of white-box attacks in the physical world.}
\label{quantitative_results_white_box}
\centering
\setlength{\tabcolsep}{1.72mm}
\begin{tabular*}{\hsize}{ccccccccccccccccccccc}
\hline\hline
ADs &PAs &P1 &P2   &P3  &P4  &P5  &P6  &P7  &P8  &P9  &P10  &P11  &P12  &P13  &P14 &P15 &P16 &P17 &P18 &Avg  \\\hline\hline
\multirow{4}{*}{AD1} &Clean&0.92&0.91&0.90&0.89&0.91&0.89&0.89&0.89&0.91&0.91&0.92&0.90&0.92&0.91&0.89&0.91&0.90&0.90&0.904   \\
&PA1&0.82&0.77&0.75&0.50&0.83&0.00&0.84&0.84&0.87&0.73&0.80&0.86&0.82&0.82&0.60&0.83&0.87&0.87&0.746    \\
&PA2&0.00&0.81&0.62&0.62&0.88&0.00&0.82&0.85&0.85&0.82&0.87&0.84&0.77&0.76&0.61&0.83&0.86&0.86&0.704  \\
&Ours&\textbf{0.00}&\textbf{0.00}&\textbf{0.00}&\textbf{0.00}&\textbf{0.00}&\textbf{0.00}&\textbf{0.00}&\textbf{0.00}&\textbf{0.00}&\textbf{0.00}&\textbf{0.00}&\textbf{0.00}&\textbf{0.00}&\textbf{0.00}&\textbf{0.00}&\textbf{0.00}&\textbf{0.00}&\textbf{0.00}&\textbf{0.000}  \\\hline

\multirow{4}{*}{AD2} 
&Clean&1.00&1.00&1.00&1.00&1.00&0.93&1.00&1.00&1.00&1.00&1.00&1.00&1.00&1.00&1.00&1.00&1.00&1.00&0.996   \\
&PA1&\textbf{0.00}&0.99&0.97&0.21&0.37&0.00&1.00&1.00&0.98&0.76&0.95&1.00&0.98&\textbf{0.36}&0.96&1.00&1.00&1.00&0.752    \\
&PA2&\textbf{0.00}&0.90&0.00&\textbf{0.00}&\textbf{0.00}&0.00&0.99&0.99&0.99&\textbf{0.00}&\textbf{0.66}&1.00&0.55&0.72&0.27&1.00&0.00&0.99&0.503  \\
&Ours&0.27&\textbf{0.33}&\textbf{0.00}&0.93&0.99&\textbf{0.00}&\textbf{0.00}&\textbf{0.00}&\textbf{0.00}&0.32&0.98&\textbf{0.00}&\textbf{0.00}&0.93&\textbf{0.00}&\textbf{0.00}&\textbf{0.00}&\textbf{0.00}&\textbf{0.264} \\\hline

\multirow{4}{*}{AD3}
&Clean&1.00&1.00&1.00&1.00&1.00&0.99&1.00&1.00&1.00&1.00& 1.00&1.00&1.00&1.00&0.97&1.00&1.00&1.00&0.998   \\
&PA1&0.97&1.00&1.00&0.95&1.00&0.86&1.00&1.00&1.00&1.00&1.00&1.00&1.00&1.00&0.91&0.99&1.00&1.00&0.982    \\
&PA2&0.98&1.00&0.93&0.25&1.00&0.74&0.99&1.00&1.00&0.84&0.94&0.97&0.97&1.00&0.00&1.00&1.00&1.00&0.867  \\
&Ours&\textbf{0.00}&\textbf{0.40}&\textbf{0.00}&\textbf{0.00}&\textbf{0.00}&\textbf{0.00}&\textbf{0.00}&\textbf{0.00}&\textbf{0.00}&\textbf{0.00}&\textbf{0.00}&\textbf{0.00}&\textbf{0.00}&\textbf{0.00}&\textbf{0.00}&\textbf{0.00}&\textbf{0.00}&\textbf{0.00}&\textbf{0.022}   \\\hline

\multirow{4}{*}{AD4} 
&Clean&1.00&1.00&1.00&1.00&1.00&0.79&1.00&1.00&1.00&1.00&1.00&1.00&1.00&1.00&1.00&1.00&1.00&1.00&0.988   \\
&PA1&0.22&0.89&1.00&0.47&0.91&0.00&1.00&1.00&1.00&0.97&1.00&1.00&0.98&1.00&0.49&0.99&0.99&0.97&0.827    \\
&PA2&1.00&0.91&1.00&0.82&0.94&0.00&1.00&1.00&1.00&1.00&1.00&1.00&1.00&1.00&0.43&0.99&1.00&1.00&0.894  \\
&Ours&\textbf{0.00}&\textbf{0.00}&\textbf{0.00}&\textbf{0.00}&\textbf{0.00}&\textbf{0.00}&\textbf{0.00}&\textbf{0.00}&\textbf{0.00}&\textbf{0.00}&\textbf{0.00}&\textbf{0.00}&\textbf{0.00}&\textbf{0.00}&\textbf{0.00}&\textbf{0.00}&\textbf{0.00}&\textbf{0.00}&\textbf{0.000}   \\
\hline\hline
\end{tabular*}
    \begin{tablenotes}
        \footnotesize
        \item {The best attack performance are highlighted in \textbf{bold}. }
    \end{tablenotes}
\end{table*}

\begin{table*}[t!]
\footnotesize
\caption{Quantitative experimental results of black-box attacks in the physical world.}
\label{quantitative_results_black_box}
\centering
\setlength{\tabcolsep}{1.74mm}
\begin{tabular*}{\hsize}{ccccccccccccccccccccc}
\hline\hline
ADs &PAs &P1 &P2   &P3  &P4  &P5  &P6  &P7  &P8  &P9  &P10  &P11  &P12  &P13  &P14 &P15 &P16 &P17 &P18 &Avg  \\\hline\hline
\multirow{3}{*}{AD2}
&PA1&0.99&0.94&0.99&0.89&0.78&0.00&1.00&1.00&1.00&0.95&0.99&1.00&0.96&0.63&0.98&1.00&0.99&0.99&0.893    \\
&PA2&0.00&0.99&0.00&\textbf{0.00}&0.82&0.00&0.99&1.00&0.98&\textbf{0.44}&0.92&1.00&0.94&\textbf{0.25}&0.33&0.99&0.54&0.99&0.609  \\
&Ours&\textbf{0.00}&\textbf{0.24}&\textbf{0.00}&0.51&\textbf{0.27}&\textbf{0.00}&\textbf{0.00}&\textbf{0.00}&\textbf{0.00}&0.80&\textbf{0.00}&\textbf{0.00}&\textbf{0.00}&0.49&\textbf{0.00}&\textbf{0.00}&\textbf{0.00}&\textbf{0.00}&\textbf{0.128}   \\\hline

\multirow{3}{*}{AD3}
&PA1&1.00&1.00&1.00&0.99&1.00&0.42&1.00&1.00&1.00&1.00&1.00&1.00&1.00&1.00&0.99&1.00&1.00&1.00&0.967    \\
&PA2&0.98&1.00&1.00&1.00&1.00&0.95&1.00&1.00&1.00&0.95&0.99&1.00&1.00&1.00&1.00&1.00&1.00&1.00&0.993  \\
&Ours&\textbf{0.00}&\textbf{0.33}&\textbf{0.00}&\textbf{0.00}&\textbf{0.00}&\textbf{0.00}&\textbf{0.00}&\textbf{0.00}&\textbf{0.00}&\textbf{0.00}&\textbf{0.00}&\textbf{0.00}&\textbf{0.00}&\textbf{0.20}&\textbf{0.00}&\textbf{0.00}&\textbf{0.00}&\textbf{0.00}&\textbf{0.029}   \\\hline

\multirow{3}{*}{AD4}
&PA1&1.00&0.99&1.00&0.94&0.73&0.00&1.00&1.00&1.00&0.99&1.00&1.00&0.85&0.99&0.99&1.00&0.98&1.00&0.914    \\
&PA2&0.98&1.00&1.00&1.00&1.00&0.00&1.00&1.00&1.00&0.85&1.00&1.00&0.96&1.00&0.88&1.00&0.92&0.99&0.921  \\
&Ours&\textbf{0.00}&\textbf{0.00}&\textbf{0.00}&\textbf{0.40}&\textbf{0.21}&\textbf{0.00}&\textbf{0.00}&\textbf{0.00}&\textbf{0.00}&\textbf{0.00}&\textbf{0.00}&\textbf{0.00}&\textbf{0.00}&\textbf{0.00}&\textbf{0.00}&\textbf{0.00}&\textbf{0.00}&\textbf{0.00}&\textbf{0.034}   \\
\hline\hline
\end{tabular*}
    \begin{tablenotes}
        \footnotesize
        \item {The best attack performance are highlighted in \textbf{bold}. The proxy model is YOLOv5 (AD1).}
    \end{tablenotes}
\end{table*}

\subsection{Contextual Attacks}

\begin{figure}[!t]
  \centering
  \includegraphics[width=0.98\linewidth]{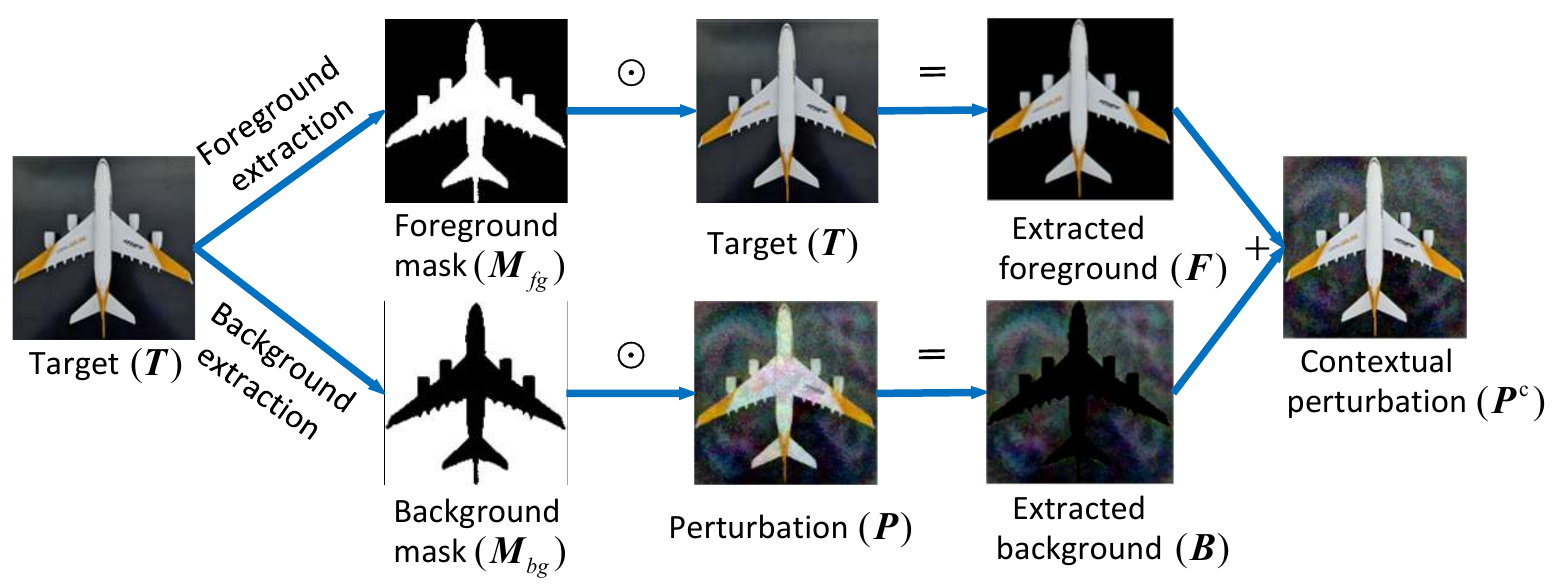}
  \caption{Contextual perturbation design.}
  \label{fig:Patch_design}
\end{figure}

Our proposed contextual attacks framework is mainly inspired by the following observations:
\begin{itemize}
    \item Contextual features matters in aerial detection.
    \item Bigger contextual perturbations with stronger attack performance.
    \item Closer distance between perturbations and targets with stronger attack performance.
\end{itemize}
Therefore, to achieve powerful uncovered attacks in physical scenarios, we propose to manipulate the contextual area of the interested targets to elaborate contextual perturbations.

Specifically, we first extract the masks of the foreground $\boldsymbol{M}_{fg}$ and background $\boldsymbol{M}_{bg}$ of the targeted object $\boldsymbol{T}$.
Secondly, contextual perturbation's foreground $\boldsymbol{F}$ and background $\boldsymbol{B}$ area are extracted from the interested target $\boldsymbol{T}$ and updated perturbation $\boldsymbol{P}$ respectively.
Thirdly, $\boldsymbol{F}$ and $\boldsymbol{B}$ are combined to formulate the contextual perturbation.
Finally, repeat the above steps until the end of training.
Mathematically, contextual perturbation is defined as:
\begin{equation}
    \label{eq:contextual_perturbation}
    \begin{aligned}
        \boldsymbol{P}^{c} & = \boldsymbol{T} \odot \boldsymbol{M}_{fg} + \boldsymbol{P} \odot \boldsymbol{M}_{bg}\\
                           & = \boldsymbol{F} + \boldsymbol{B}
    \end{aligned}
\end{equation}

\subsection{Loss Function}

To fool detectors in physical scenarios, the loss function consists of two components, adversary loss and smoothness loss.

\textbf{Adversary loss}:
We use objectiveness scores of all detected objects to optimize the contextual perturbations. 
Therefore, the adversarial loss is written as:
\begin{equation}
    \label{eq:Lobj}
    L_{adv} = \frac{1}{n} \sum\limits_{i=1}^n P_i(obj),
\end{equation}
where $P_i(obj)$ means the objectiveness score of $i$th detected interested target , and $n$ represents the number of detected interested targets. 
The adversarial loss is adopted to gift the contextual perturbations with attack efficacy during training.

\textbf{Smoothness loss}:
Existing works demonstrate that perturbation's smoothness is crucial in maintaining attack efficacy during imaging.
Since imaging devices can barely capture the value gap between adjacent pixels, total variation (TV) \cite{sharif2016accessorize} is adopted as the smoothness limitation of the generated adversarial perturbations.
TV can be written as: 
\begin{equation}
    \label{eq_Ltv}
    L_{tv} = \sum\limits_{i,j} \sqrt{(p_{i+1,j}-p_{i,j})^2 + (p_{i,j+1}-p_{i,j})^2},
\end{equation}
where $p_{i,j}$ represents the pixel value of $i$th row, $j$th column of the optimized adversarial perturbation.

Consequently, the total loss function is as follows:
\begin{equation}
    \label{eq:total_loss}
    L = L_{adv} + \lambda \cdot L_{tv},
\end{equation}
where $\lambda$ is used to balance the two parts of the total loss. 

\section{EXPERIMENTS}
\label{sec:experiments}

\subsection{Experimental Settings}

In experiments, public datasets DOTA \cite{xia2018dota} and RSOD\footnote{\url{https://github.com/RSIA-LIESMARS-WHU/RSOD-Dataset-}} are used to train aerial detectors (ADs) and contextual perturbations, respectively.
Moreover, we choose various object detection methods to verify the attack effectiveness of the proposed method, including YOLOv5\footnote{\url{https://github.com/ultralytics/yolov5}} (\textbf{AD1}), Faster R-CNN \cite{ren2015faster} (\textbf{AD2}), Swin Transformer \cite{liu2021swin} (\textbf{AD3}), FreeAnchor \cite{zhang2019freeanchor} (\textbf{AD4}).
Two SOTA physical attacks (PAs) are chosen for comparison, including the adversarial perturbations generated by Thys \etal \cite{thys2019fooling} (\textbf{PA1}) and APPA  \cite{lian2022benchmarking} (\textbf{PA2}). 

\subsection{Experimental Results}

\begin{figure}
  \centering
  \begin{subfigure}{0.49\linewidth}
    \includegraphics[width=1\linewidth]{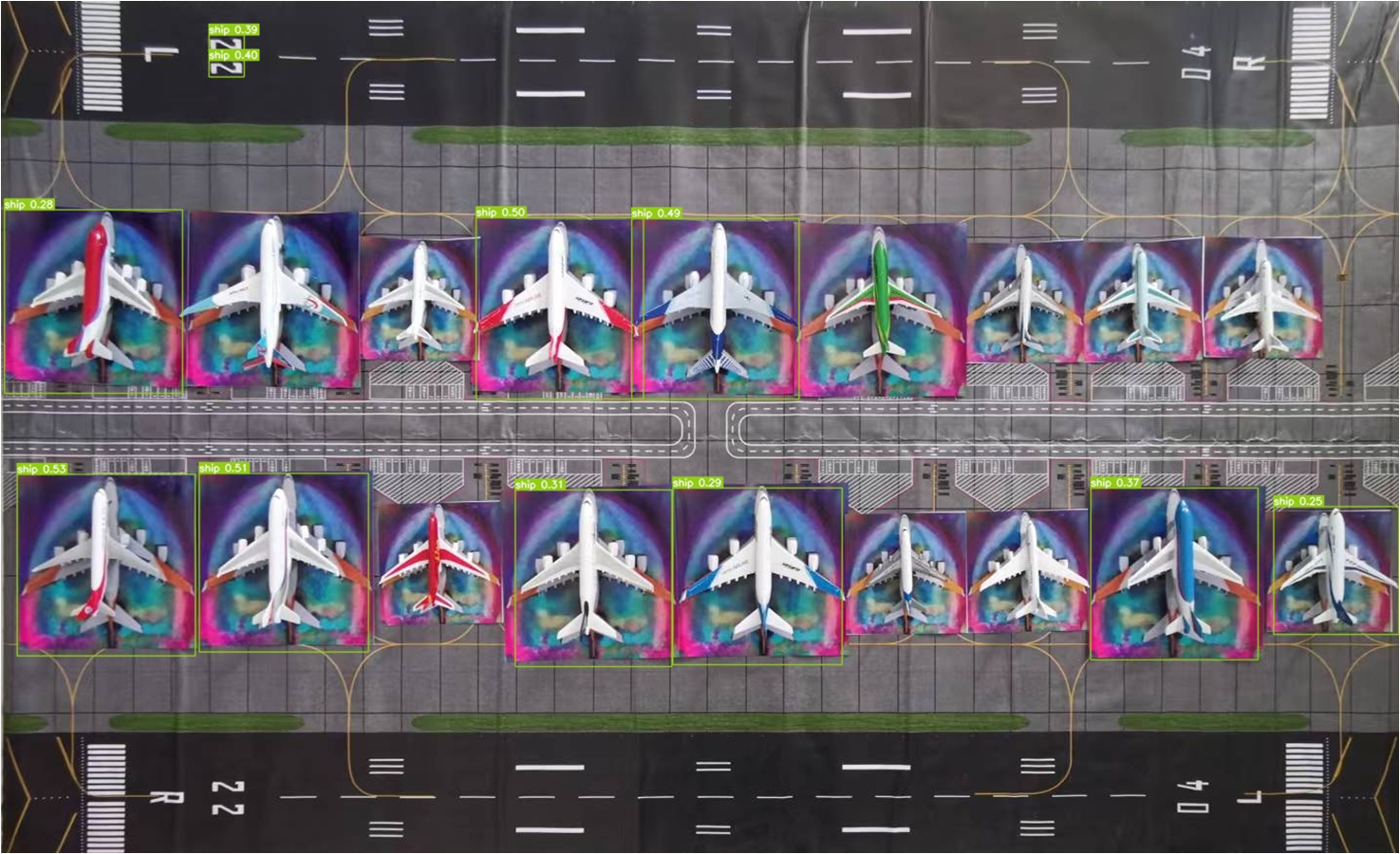}
    \caption{YOLOv5}
  \end{subfigure}
  \begin{subfigure}{0.49\linewidth}
    \includegraphics[width=1\linewidth]{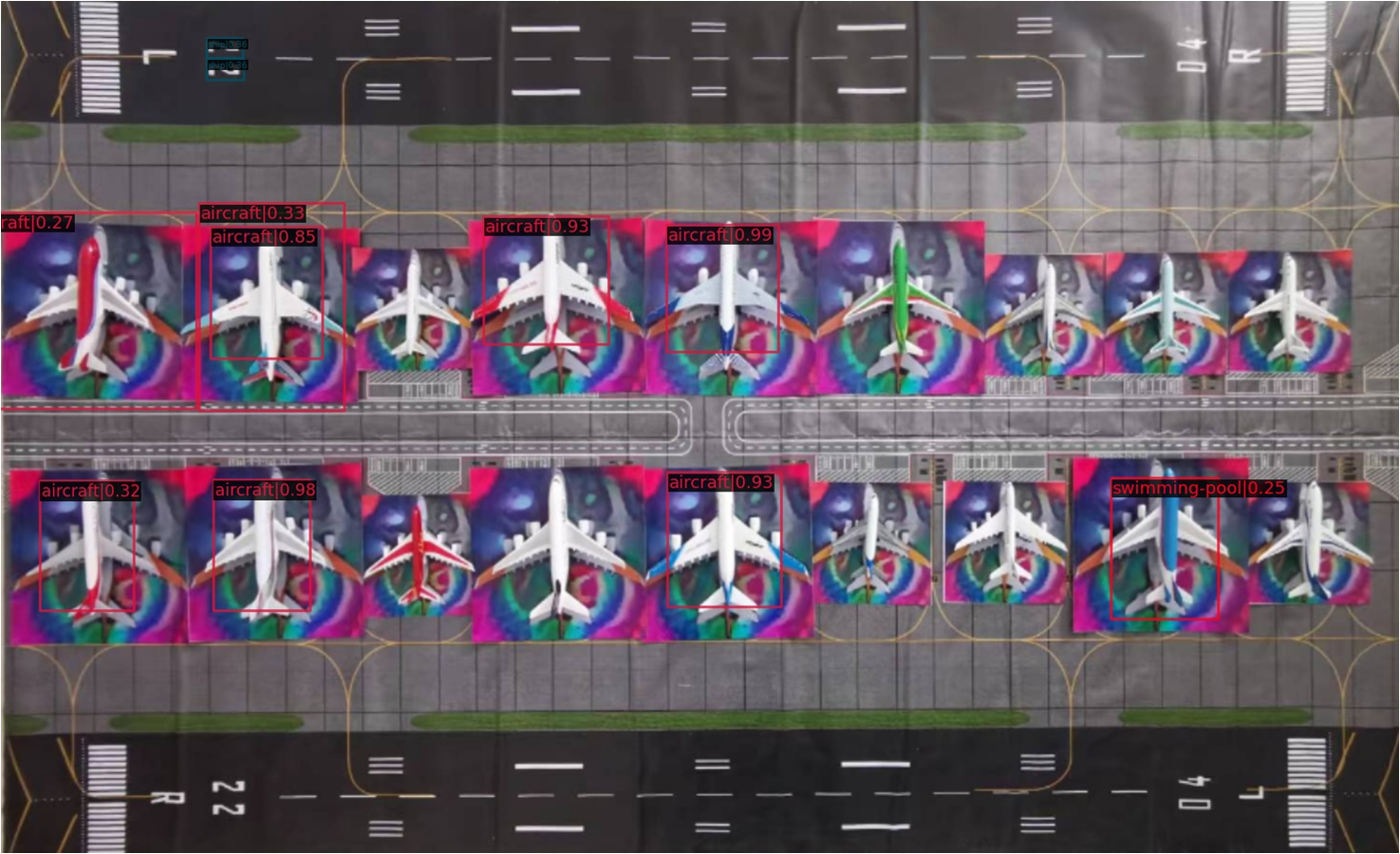}
    \caption{Faster R-CNN}
  \end{subfigure}
  \begin{subfigure}{0.49\linewidth}
    \includegraphics[width=1\linewidth]{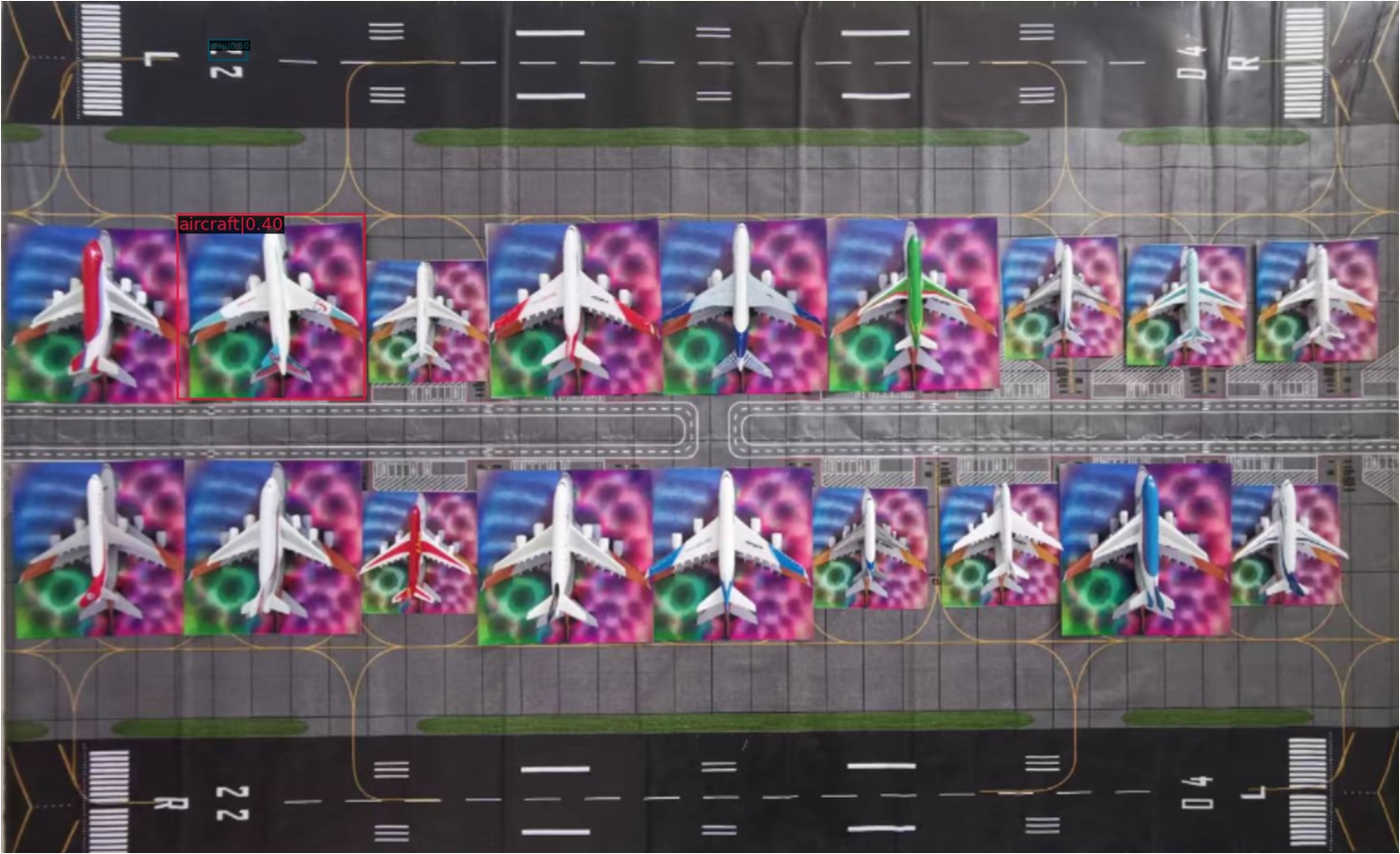}
    \caption{Swin Transformer}
  \end{subfigure}
  \begin{subfigure}{0.49\linewidth}
    \includegraphics[width=1\linewidth]{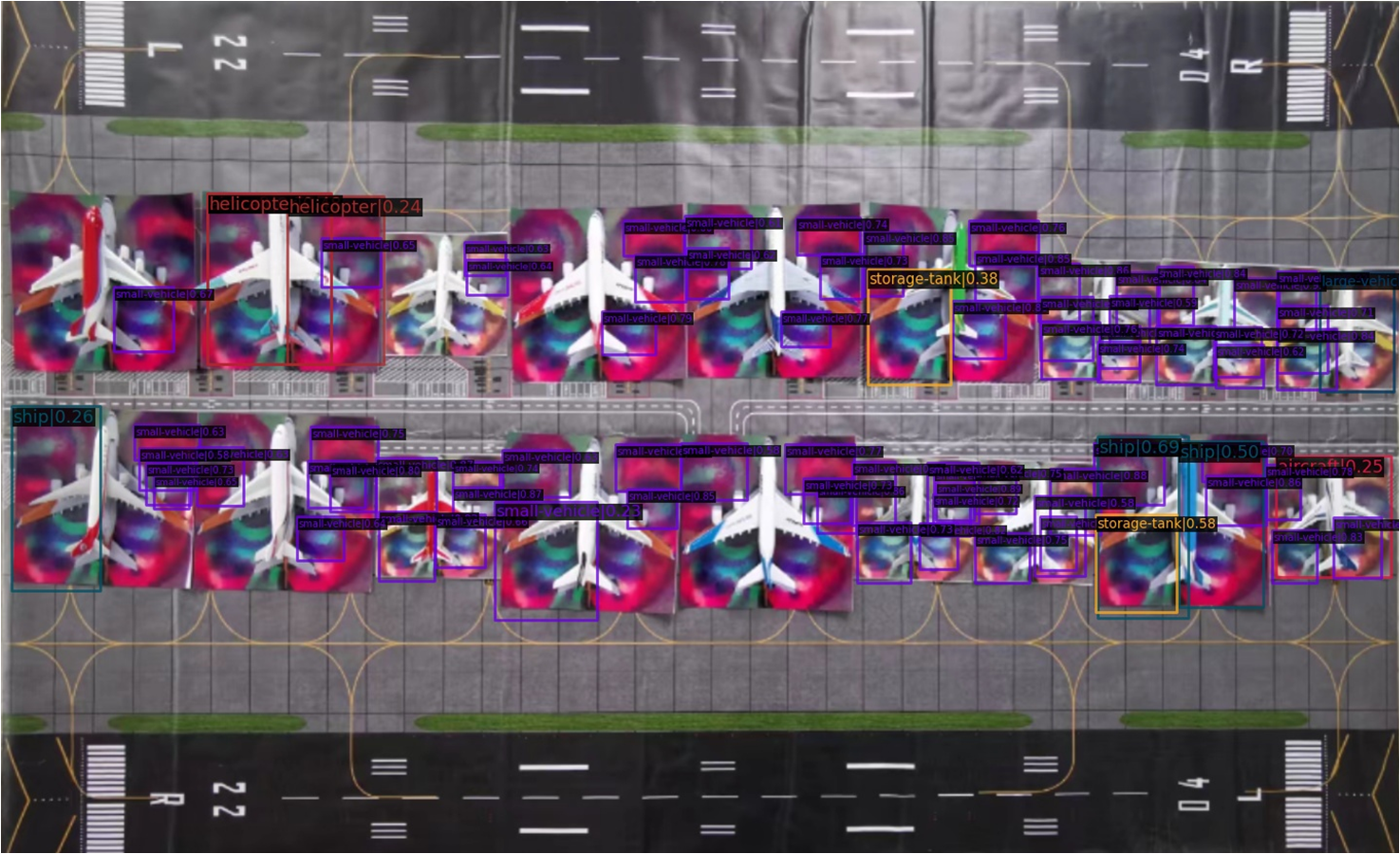}
    \caption{FreeAnchor}
  \end{subfigure}
  \caption{Qualitative attack performance in the physical world.}
  \label{fig:qualitative_results}
\end{figure}

The planes are chosen as the interested targets in the proportionally scaled experiments.
We use the detection confidence scores of 18 plane models (\textbf{P1}-\textbf{P18}) to compare the physical attack performance, \ie, the lower the confidence scores, the better the attack performance.

The quantitative experimental results of white-box and black-box attacks (YOLOv5 is selected as a proxy model to train contextual perturbations) are shown in Table \ref{quantitative_results_white_box} and Table \ref{quantitative_results_black_box} respectively.
It is observed that our method shows great superiority in both attack efficacy and transferability. 
Specifically, our elaborated contextual perturbations can significantly lower the average detection confidence of all aerial detectors, from 0.904, 0.996, 0.998, 0.988 to 0.000, 0.264, 0.022, 0.000, far better than the comparison methods.
Moreover, our method can drop the average detection confidence from 0.893, 0.967, 0.914 to 0.128, 0.029, 0.034, even in black-box settings.
The qualitative experimental results are shown in Fig \ref{fig:qualitative_results}. 
We can observe that the generated contextual perturbations of our proposed contextual attack method can easily blind various aerial detectors, even after digital-physical domain transformation.

\section{CONCLUSION}
\label{sec:conclusion}

In this article, we aim to hide interested targets from being detected by various aerial detectors without smearing targeted objects.
To achieve that, we propose a novel contextual attack against aerial detection in physical world scenarios, which fully uses the interested targets' contextual features to elaborate contextual perturbations and achieves the best attack performance in both white-box and black-box settings.
Extensive experiments demonstrate the effectiveness and superiority of our proposed contextual attack method.

% References should be produced using the BibTeX program from suitable
% BibTeX files (here: strings, refs, manuals). The IEEEbib.bst bibliography
% style file from IEEE produces an unsorted bibliography list.
% -------------------------------------------------------------------------
\bibliographystyle{IEEEbib}
\bibliography{refs}

\end{document}